\pdfoutput=1

\documentclass[11pt]{article}

\usepackage{acl}

\usepackage{times}
\usepackage{latexsym}
\usepackage{subcaption}
\usepackage{graphics}
\usepackage{graphicx}
\usepackage{hyperref}
\usepackage{dsfont}
\usepackage{amsmath}
\usepackage{xspace}
\usepackage{verbatim}
\usepackage{xcolor}
\usepackage{soul}
\usepackage{graphicx}
\usepackage{tabularx}
\usepackage{booktabs}
\usepackage{caption}
\usepackage{geometry}
\usepackage{caption}
\usepackage{comment}
\usepackage{multirow}

\usepackage[T1]{fontenc}

\usepackage[utf8]{inputenc}

\usepackage{microtype}

\usepackage{todonotes}

\makeatletter
\def\blfootnote{\xdef\@thefnmark{}\@footnotetext}
\makeatother

\title{Defending Against Disinformation Attacks in \\ Open-Domain Question Answering
}

\author{Orion Weller*\ \ \ \  \ \ Aleem Khan*\\
\textbf{Nathaniel Weir\ \ \ \ \ \  Dawn Lawrie\ \ \ \ \ \ Benjamin Van Durme} \\ 
Johns Hopkins University \\
\normalsize{\texttt{\{oweller,aleem\}@cs.jhu.edu}}
}

\begin{document}
\maketitle
\begin{abstract}
Recent work in open-domain question answering (ODQA) has shown that adversarial poisoning of the search collection can cause large drops in accuracy for production systems. 
However, little to no work has proposed methods to defend against these attacks.
To do so, we rely on the intuition that redundant information often exists in large corpora. 
To find it, we introduce a method that uses query augmentation to search for a diverse set of passages that could answer the original question but are less likely to have been poisoned.
We integrate these new passages into the model through the design of a novel confidence method, comparing the predicted answer to its appearance in the retrieved contexts (what we call \textit{Confidence from Answer Redundancy}, i.e. CAR).
Together these methods allow for a simple but effective way to defend against poisoning attacks that provides gains of nearly 20\% exact match across varying levels of data poisoning/knowledge conflicts.\footnote{Code and data will be made public at \url{https://github.com/orionw/disinformation-defense}}
\blfootnote{* Authors contributed equally}
\end{abstract}

\section{Introduction}
\label{sec:intro}
Open-domain question answering (ODQA) is the task of answering a given question based on evidence from a large corpus of documents. In order to do so, a system generally first retrieves a smaller subset of documents (typically between 5-100) and then answers the question based on those documents. Previous research in ODQA has resulted in many well-curated datasets that evaluate a model's ability to answer questions on a wide array of topics~\cite{kwiatkowski2019natural,joshi2017triviaqa, Dunn2017SearchQAAN,yang2015wikiqa}.

\begin{figure}[!t]
    \centering
    \includegraphics[width=\columnwidth, trim={0.25cm 0cm 2mm 0cm}]{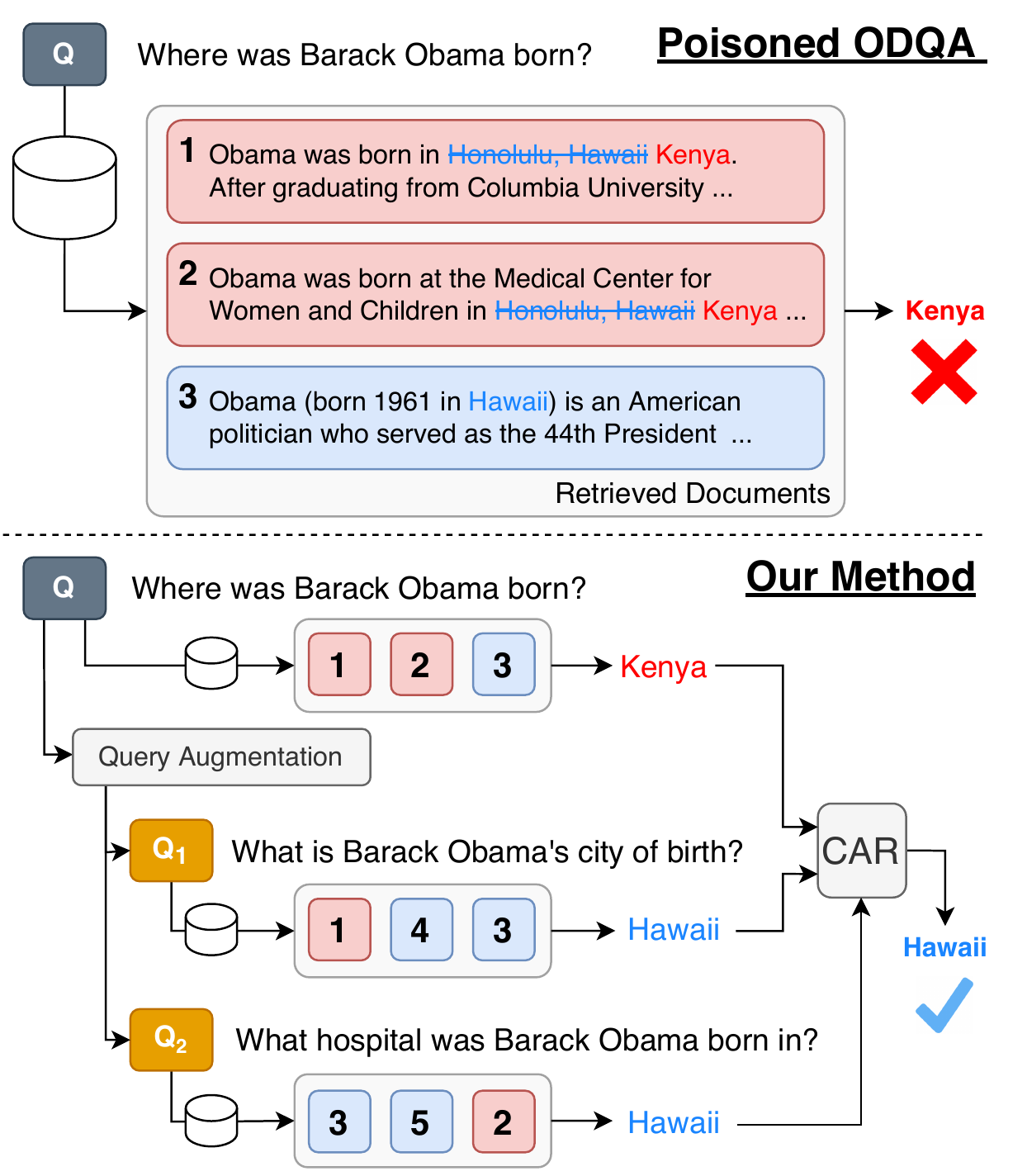}
    \caption{An example of a poisoning attack on an open-domain question answering (ODQA) pipeline with our method (Lower) vs a standard system (Upper).
    The passages have been adversarially poisoned to change Obama's \textcolor[RGB]{26,133,255}{correct} birthplace to be \textcolor[RGB]{255,0,0}{incorrect}. 
    Our proposed defense method uses query augmentation to find new contexts that are less likely to be poisoned (\textcolor[RGB]{26,133,255}{\#4} and \textcolor[RGB]{26,133,255}{\#5}). It then uses a novel confidence-based aggregation method (CAR) to predict the correct answer.
    \vspace{-1em}}
    \label{fig:example}
\end{figure}

However, most internet users search across less-carefully curated sources, where malicious actors are able to affect articles that may be used by an ODQA system (\autoref{fig:example}). Furthermore, even in curated knowledge sources like Wikipedia, we frequently see attacks (e.g. malicious edits/fake pages) that have even impacted production QA systems.\footnote{For examples of disinformation attacks on popular entities that motivate our approach see Appendix~\ref{app:popularity} or the \href{https://en.wikipedia.org/wiki/Reliability_of_Wikipedia\#Notable_incidents}{``Reliability of Wikipedia"} or \href{https://en.wikipedia.org/wiki/Vandalism_on_Wikipedia}{``Vandalism on Wikipedia"} pages.}

Recent work has recognized the potential for bad actors to influence automated knowledge-intensive NLP systems that involve retrieval: \citet{du2022synthetic} explored how poisoned information affects automated fact verification systems using sparse non-neural information retrieval systems, while \citet{Chen2022RichKS,longpre2021entity,pan2023risk} have studied the effect of knowledge conflicts and poisoning attacks on ODQA pipelines. All of these works have illustrated that poisoning attacks significantly decrease system performance, even when using state-of-the-art models; however, only \citet{pan2023risk} has even briefly considered the task of \textit{defending} against poisoning attacks (which are becoming increasingly common, see Appendix~\ref{app:popularity} for real-life examples) and their proposed method, majority voting over different documents, provides only minor gains.

We seek to fill this gap by proposing a simple but effective defense against these attacks. Building on the intuition that information is usually available in multiple places and that it is unlikely that all sources (or pages) will be poisoned, we propose a novel query augmentation scheme to gather a larger set of diverse passages. We also propose a new confidence method to decide when to use the newly gathered contexts vs the original, which we call \textit{Confidence from Answer Redundancy} (CAR).

Our proposed approach involves no gradient updates, can easily be applied to existing frameworks, and uses a simple resolution approach to arrive at the predicted answer. Together, our methods can provide gains of nearly 20 points in exact match, helping to reduce the negative effects of data poisoning and disinformation attacks on ODQA.

\section{Experimental Details}
\label{sec:exp}
We seek to mimic realistic disinformation attacks on a curated knowledge source; thus, for our experiments we use Wikipedia as the knowledge collection for both original and augmented queries, and simulate an attack on each question independently. We follow \citet{du2022synthetic} and poison the entirety of each Wikipedia page that corresponds to each of the retrieved passages.\footnote{e.g. if at least one of the 100 retrieved passages was from Obama's Wikipedia page, the rest of his page is poisoned} We vary the amount of poisoned pages from 1 to 100.\footnote{As 100 passages are given to the models (so 100 is all passages - see Appendix~\ref{app:not_zero} for why scores are non-zero). We also experimented with poisoning random retrieved passages in the top 100 and found similar results (Appendix~\ref{app:poisoned_contexts})}  Note that we do not poison the entire corpus, as poisoning millions of pages is beyond the scope of common attacks.

\subsection{Data}
For our experiments we use Natural Questions (NQ) \cite{kwiatkowski2019natural} and TriviaQA \cite{joshi2017triviaqa}, two popular datasets for open-domain question answering. Furthermore, previous research on conflicts in ODQA has used these datasets in their experiments \cite{Chen2022RichKS}. The Natural Question dataset was gathered by collecting real-user queries typed into Google Search, while TriviaQA was collected by scraping question and answer pairs from trivia websites, and then matching the answers to Wikipedia passages.

We simulate the data poisoning through the code available from \citet{longpre2021entity}, which introduced the problem in ODQA and has been used in subsequent work \cite{Chen2022RichKS}. Their method uses the answers to the questions to suggest an entity of the same type, using SpaCY NER \cite{spacy2}, which is then used to replace the correct answer in the text. This allows for entity substitutions that keep the semantic order of the context, such as replacing dates with dates, people with people, numbers with numbers, etc.

\subsection{Models}
We use two SOTA models: Fusion-in-Decoder (FiD) and \textsc{Atlas}. FiD is an encoder-decoder architecture that generates an answer by first retrieving and encoding $N$ passages and then concatenating them and giving them to the decoder \cite{Izacard2021LeveragingPR}. FiD uses DPR for retrieval \cite{karpukhin2020dense}. \textsc{Atlas}  \cite{izacard2022few} is currently the state-of-the-art model on Natural Questions and TriviaQA. This model also uses fusion in the decoder and has a T5 backbone, but uses Contriever \cite{izacard2022unsupervised} for retrieval and does joint end-to-end training. For information on hyperparameters see Appendix~\ref{app:hyperparameters}.

\section{Proposed Method}\label{sec:propose}

\subsection{Query Augmentation}
\label{sec:gen}
We hypothesize that in cases of conflicting evidence in large corpora for \emph{factoid} based questions, there will generally be more evidence for the correct answer than for incorrect ones. For example, imagine the question ``Where was Barack Obama born?" with a corresponding attack to his Wikipedia page (see \autoref{fig:example}). As Wikipedia contains redundant information, alternate questions that find contexts on other pages (e.g. his mother \textit{Ann Dunham}'s page) will still find the right answer.

\begin{figure}[t]
\centering
Natural Questions \\
\includegraphics[scale=.8,trim={1.5cm 0.25cm 1cm 0.05cm}]{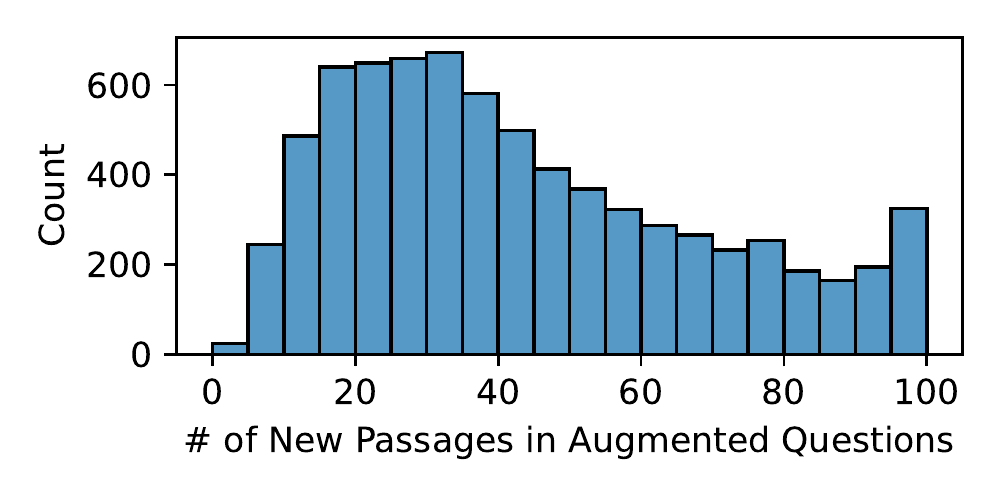} \\
TriviaQA \\
\includegraphics[scale=.8,trim={1.5cm 0.4cm 1cm 0.05cm}]{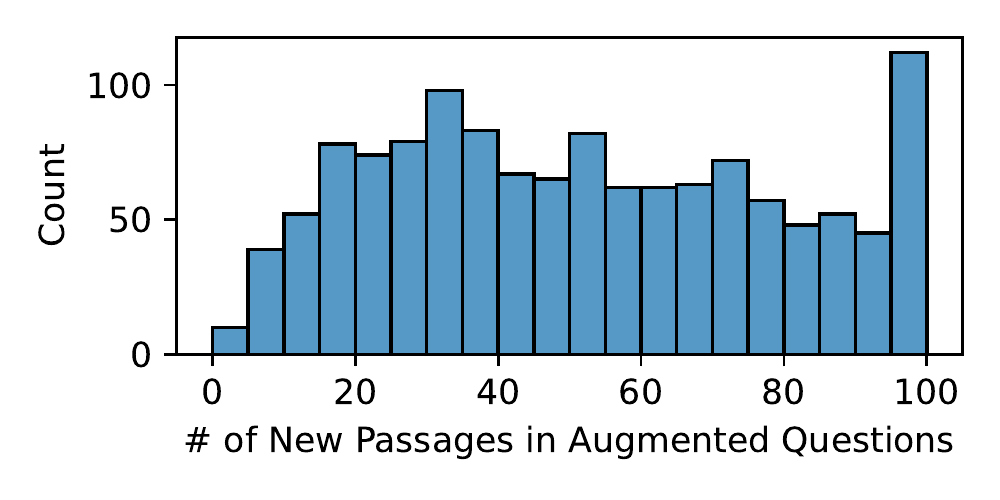}
\caption{Number of new passages retrieved per augmented question (e.g., a question in the 100 bin would have 100 new contexts not retrieved by the original). Natural Questions is on top and TriviaQA on bottom.\vspace{-0.5em}}
\label{fig:uniquehist}
\end{figure}

To create these alternate questions that will still find the correct answer but with more diverse passages, we propose a query augmentation scheme that has similarities to query expansion in information retrieval 
 (IR) \cite{singhal2001modern,carpineto2012survey,wei2022decompositions,claveau2021neural}. We generate these new questions for each original question by prompting GPT-3. We use \texttt{davinci-002} from \citet{brown2020language}, but one can alternatively use open-source language models for similar results: see \autoref{tab:gpt_vs_llama_nq} and Appendix~\ref{app:llama} for results with Vicuna v1.5 (using Llama 2). These query augmentations are not necessarily paraphrases as they strive to be as different as possible while still leading to the correct answer. They are also not identical to classic query expansion from IR either, as they do not intend to solely broaden the query scope but rather to find diverse contexts from questions of any scope.

 For each query in the dataset, we prompt GPT-3 with the following: \texttt{"Write 10 new wildly diverse questions with different words that have the same answer as \{Original Question\}"}, thus generating approximately 10 augmented questions per original question (c.f. \autoref{table:datasets} for three examples of generations). Finally, we retrieve the 100 most relevant contexts for those augmented questions. Note that if searching with the augmented questions retrieves a passage from a Wikipedia page that was already poisoned from the initial set of 100 (see Section~\ref{sec:exp}) we return the poisoned text following \citet{du2022synthetic}.

When we compare these newly retrieved passages to the passages retrieved by the original question we find that they do provide a more diverse set of passages. \autoref{fig:uniquehist} shows the distribution of new passages retrieved, with almost all retrieving at least 20 or more new passages and a substantial amount having an entirely new set of 100 passages.

\subsection{Confidence from Answer Redundancy}
In order to identify the best augmented queries with their corresponding new passages, we derive a novel method, CAR, for measuring ODQA confidence. CAR measures how often the predicted answer string occurs in the retrieved contexts (usually 100 contexts). For example, if the predicted answer appears only once in all 100 contexts, this may mean that the retriever was not able to find many documents relevant to the query, especially as popular entities (those asked about in NQ and TriviaQA)  are generally found in many articles. Overall, the more frequently the predicted answer appears in the contexts, the more likely that the retrieval was both successful and plentiful (e.g. redundant).

\begin{table}[t!]
\small
\begin{tabular}{l}
 When was the last time anyone was on the moon? \\
\midrule
When was the last time anybody walked on the moon? \\ 
When was the last manned mission to the moon?  \\
When was the last time a human was on the moon?  \\
\end{tabular}

\vspace{0.3cm}

\begin{tabular}{l}
In which year did Picasso die? \\
\midrule
When did Picasso die? \\
How old was Picasso when he died? \\
What was Picasso's cause of death? \\
\end{tabular}
\vspace{0.3cm}

\begin{tabular}{l}
What is the largest city in Turkey? \\
\midrule
What city in Turkey has the most people? \\
What is the most populous city in Turkey? \\
What is the most urbanized city in Turkey? \\
\end{tabular}

\caption{\label{table:datasets} Example question augmentations with the original question on top (see Appendix~\ref{app:case_study} for more).\vspace{-0.5em}
}
\end{table}

\begin{figure*}[t!]
\centering
\begin{subfigure}{.5\textwidth}
  \centering
  \includegraphics[width=.75\linewidth,trim={2cm 0.5cm 1cm 1cm}]{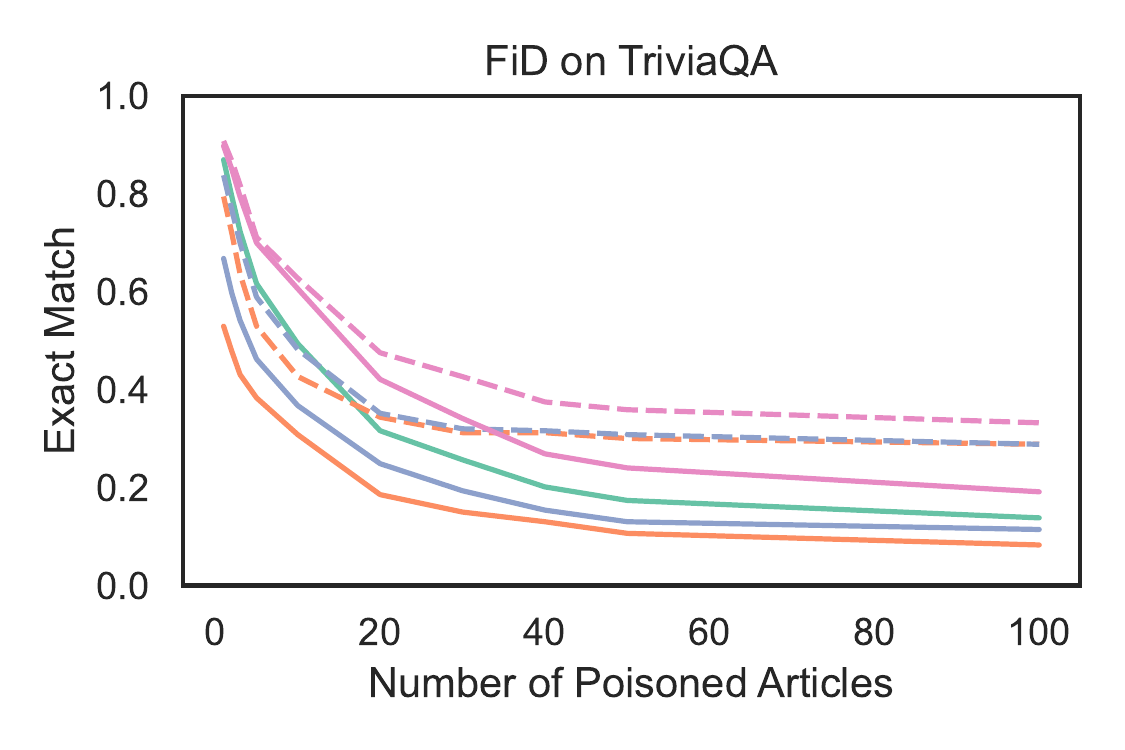}
\end{subfigure}%
\begin{subfigure}{.5\textwidth}
  \centering
  \includegraphics[width=.75\linewidth,trim={2cm 0.5cm 1cm 1cm}]{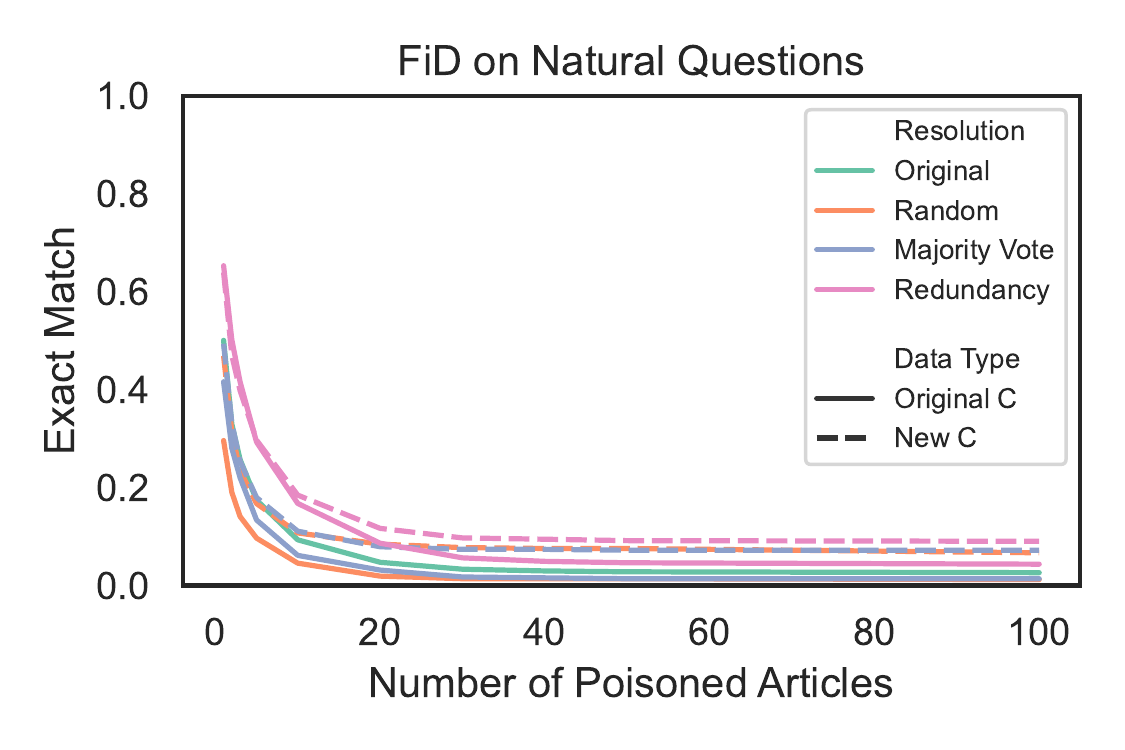}
\end{subfigure}
\vspace{-1.25em}
\end{figure*}

\begin{figure*}[htb!]
\centering
\begin{subfigure}{.5\textwidth}
  \centering
  \includegraphics[width=.75\linewidth,trim={2cm 0.5cm 1cm 1cm}]{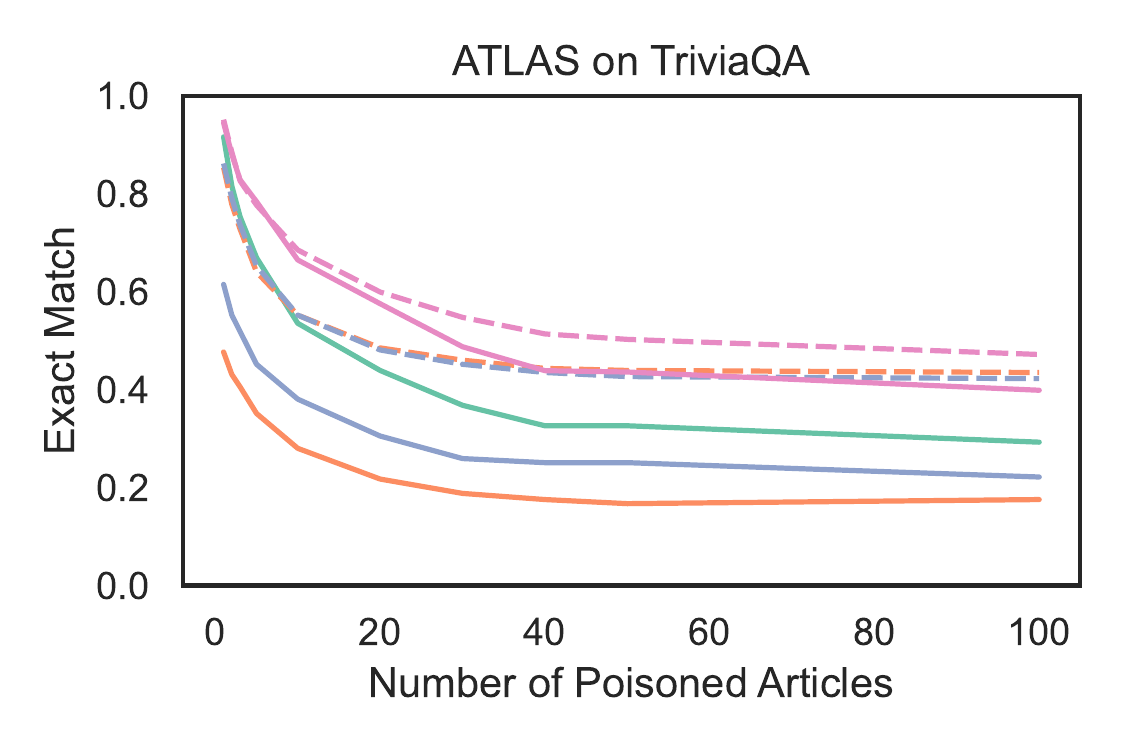}
\end{subfigure}%
\begin{subfigure}{.5\textwidth}
  \centering
  \includegraphics[width=.75\linewidth,trim={2cm 0.5cm 1cm 0cm}]{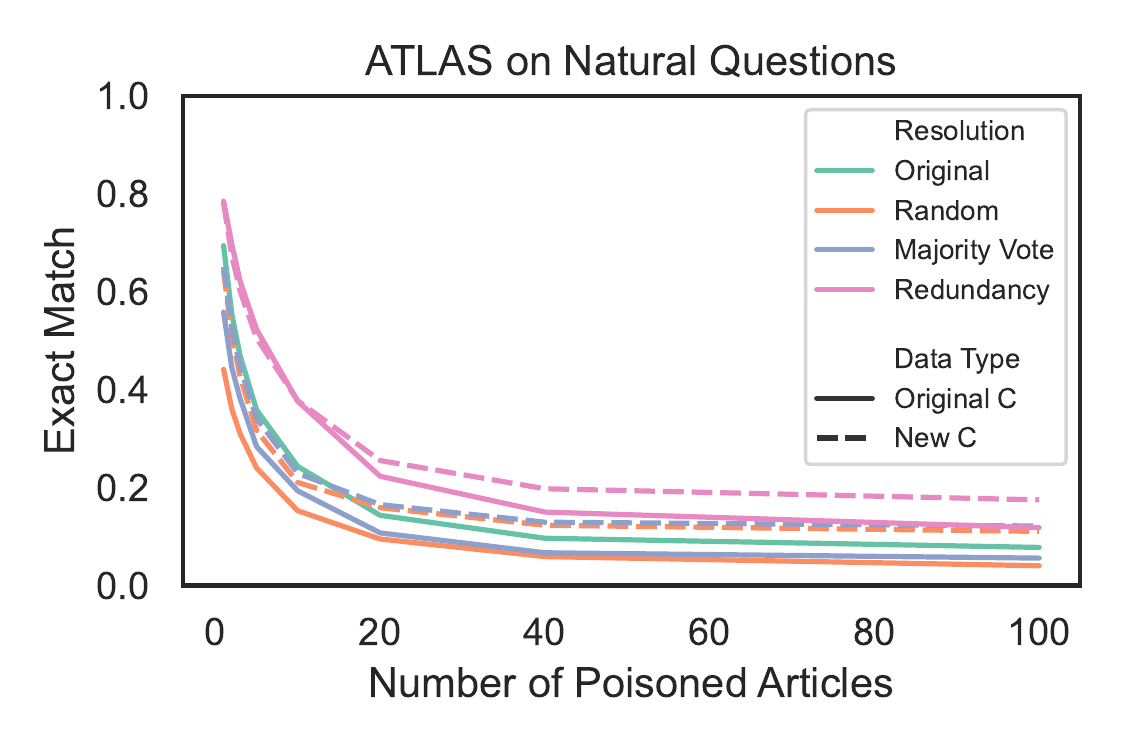}
\end{subfigure}
\caption{Data poisoning and defense strategies using \textsc{Atlas} (Lower Figure) and FiD (Upper Figure). See Appendix~\ref{app:tables} for equivalent table version of these plots. Left shows TriviaQA, right shows Natural Questions. C stands for context. 100 poisoned articles indicates all contexts are poisoned; performance is non-zero because the models ignore the contexts or the poisoning failed to recognize all aliases (\S{}\ref{app:num_poisoned}). Note that \textbf{Redundancy greatly outperforms the majority vote baseline} from  \citet{pan2023risk}. Scores plateau after around 40 poisoned articles as that is around when all 100 retrieved passages are poisoned (see Appendix~\ref{app:num_poisoned} for a discussion of article vs passage).} 
\label{fig:main}
\end{figure*}

 In practice, given a set of documents $D$, we set a hyperparameter $k$ to determine the cutoff for CAR (in practice we use $k=5$, found by tuning on the dev set). If the model retrieves more than $k$ unique passages that contain the predicted answer string, we classify the model as confident and vice versa. We use this as part of our resolution method below.

\subsection{Answer Resolution}
We use the following methods to combine (or not combine) the original question with the augmented questions, with shortened names in italics. Note that methods one through three are baselines for our newly proposed technique: (1) use the \textit{original} question only, e.g. the ``do-nothing" baseline (2) \textit{random}ly pick one new augmented question (3) take a \textit{majority vote} of the augmented question's predictions (e.g. the method from \citet{pan2023risk}) or (4) use answer \textit{redundancy}, described in the following paragraph. We also attempted several variants of these options that underperformed and are not included for clarity (Appendix~\ref{app:more_baselines}).

Our proposed method for answer resolution, \textit{redundancy}, uses CAR to effectively combine both the original question and the new augmented questions. We use CAR to decide whether to choose the original question's prediction, and if not, use a majority vote over the predictions from the augmented questions that are confident (filtered using CAR). By doing so, we retain performance from the original question and passage set when confident, while otherwise backing off to the augmentation. 
 
All methods except the baseline can use either the original (\textit{Original C}) or new (\textit{New C}) sets of passages as context and we show both options in our results. Further, majority vote and redundancy can choose between either the new or original \textit{questions} during inference (we use original, after tuning, see \autoref{app:hyperparameters} for more details).

\section{Results}\label{sec:results}
Figure \ref{fig:main} highlights our key findings using FiD and \textsc{Atlas} (for results in table form, see Appendix~\ref{app:tables}). Following \cite{longpre2021entity,Chen2022RichKS}, all results are filtered by those that the model originally predicted correctly, thus making the original method have by definition 100\% EM at the 0-article poisoning level. We show results in EM, as is typically done in previous work \cite{Izacard2021LeveragingPR,izacard2022few}, however, F1 results are nearly identical and can be found in Appendix~\ref{app:f1}.

\begin{table*}[t!]
\small
\centering
\begin{tabular}{ll|rrrrrrrrrr}
\toprule
& & \multicolumn{10}{p{8cm}}{\centering{Number of Poisoned Articles}} \\
Context Type & Resolution & 1 & 2 & 3 &  5 & 10 & 20 & 40 & 50 & 100 \\ 
\midrule
 \multirow{4}{*}{Original C} & Majority Vote & -0.6 & -0.8 & 1.0 & -0.7 & -0.4 & 0.2 & 0.0 & 0.0 & 0.0 \\
&      Original &  1.0 & -0.1 & 1.4 & 1.8 & 1.1 & 1.1 & 1.0 & 0.9 & 0.8 \\
 &        Random & -5.6 & -5.6 & -4.9 & -2.7 & -1.9 & -0.9 & -0.3 & -0.2 & -0.4  \\
&    Redundancy & 0.2 & -0.1 & 0.4 & 0.4 & 0.8 & 0.9 & 0.7 & 0.6 & 0.5 \\
 \midrule
\multirow{3}{*}{New C}  & Majority Vote & 4.7 & 3.2 & 2.8 & 2.9 & 2.3 & 1.9 & 2.5 & 2.3 & 2.3 \\
&        Random & 2.6 & 1.8 & 1.2 & 2.4 & 1.9 & 2.4 & 2.7 & 2.7 & 1.8  \\
 &    Redundancy & 1.3 & -0.4 & 1.7 & 3.4 & 2.7 & 3.0 & 3.1 & 2.9 & 2.9  \\
\bottomrule
\end{tabular}
\caption{Difference between GPT-3 and Vicuna v1.5 (using Llama 2) generations as query augmenters for NQ with FiD (positive scores indicate GPT-3 is better). Results in EM. Results are comparable to GPT-3 DaVinci in \autoref{fig:main}.\vspace{-0.75em}}
\label{tab:gpt_vs_llama_nq}
\end{table*}

As expected and shown in previous work \cite{pan2023risk,Chen2022RichKS}, we find that as the amount of poisoned data given to the model increases, performance decreases. 
We also find that resolution methods that use the new contexts (\textit{New C}) outperform those that use the original contexts, confirming the intuition behind our proposed method of finding diverse new contexts (e.g. 55.9 vs 65.1 EM for EM at 1 article poisoned).
Furthermore, we see that the \textit{redundancy} resolution strategy outperforms all other strategies (including the only published baseline, majority voting from \citet{pan2023risk}), by up to 19.4\% in the TQA setting (33.2\% at 100 poisoned articles vs 13.8\% baseline). 
Scores on NQ are lower than TQA, even with no poisoning, but still improve up to 14\% EM using redundancy.

Overall, we see that our proposed \textit{redundancy} method outperforms all other methods on both datasets, at every level of poisoning and especially so when using the newly retrieved contexts.

\paragraph{Can we use open-source LLMs as the query augmentation model?} We replace GPT-3 with Vicuna v1.5 (using Llama 2) and repeat the experiments with FiD. The results are shown in \autoref{tab:gpt_vs_llama_nq} for NQ and in Appendix~\ref{app:llama} in figure form. We see that Vicuna performs similar to GPT-3, in some cases even outperforming it. Thus, we see that our approach works with both open and closed-source models.

\begin{figure}[t!]
     \centering
     \includegraphics[scale=.75,trim={3cm 0.25cm 2cm -0.25cm}]{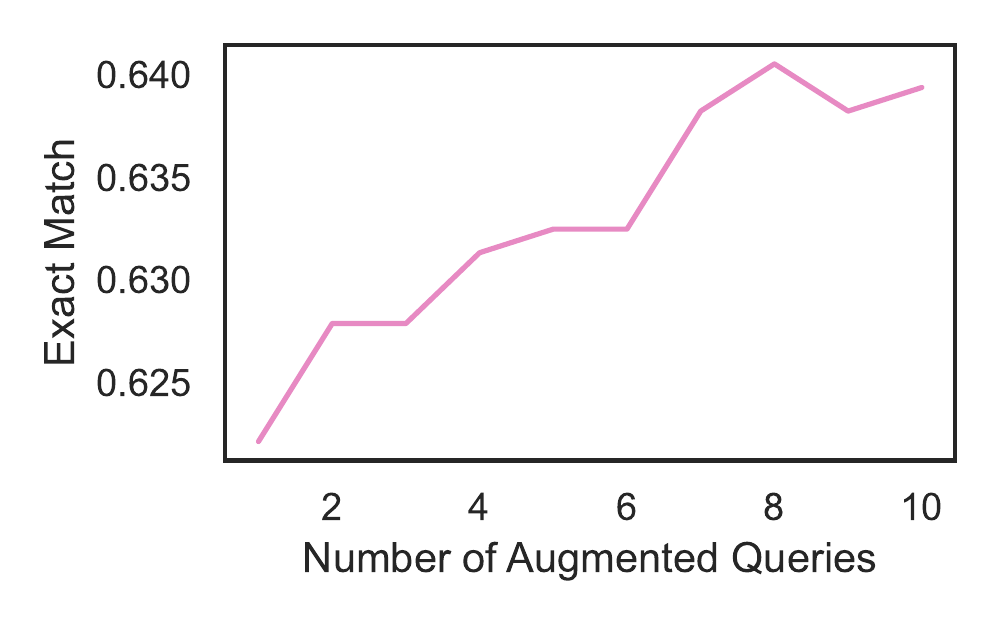}
     \caption{An ablation on the number of augmented queries (and thus number of times retrieval is used) for the \textit{redundancy} resolution method on Natural Questions 1-article FiD poisoning setting. \textbf{As the number of augmented queries increases, so does the performance}. Baseline performance is 50.1\%, indicating that even just one augmented query provides significant gains.\vspace{-0.75em}}
     \label{fig:number_queries}
\end{figure}

\paragraph{How many augmented questions are needed for our approach to work well?} To answer this, we show Figure~\ref{fig:number_queries} with the overall trend showing that as the number of augmented queries increases, so does the score. Furthermore, it shows that even one augmented query has gains over the baseline method, allowing for a more compute efficient method at the expensive of several points of performance. More computational analysis of our methods is in Appendix~\ref{app:compute}.

\paragraph{{Why is performance not 0\% at 100 poisoned documents?}} We also explore why performance is non-zero when the number of poisoned articles is equal to the number of contexts the model receives. We manually annotated 20 examples on TriviaQA that FiD got correct at the 100-article poisoning setting. We found that it is due to the model using its parametric knowledge to correctly answer (65\% of the time), as the correct answer was not present in any of the input documents, or due to answer aliases (35\%) that were not part of the answer set. Examples of cases can be found in Appendix~\ref{app:not_zero}.

\section{Conclusion}
Our work defends against data poisoning attacks in open-domain question answering through two novel methods: (1) the use of query augmentation to find diverse passages that still correctly answer the question and (2) the use of answer redundancy as a strategy for model confidence in its prediction. Our proposed methods do not involve \emph{any} gradient updates and provide a significant performance improvement. Thus, our work shows the effect of data poisoning on state-of-the-art open-domain question-answering systems and provides a way to improve poisoned performance by almost 20 points in exact match. We hope that this work encourages future work in defending against poisoning attacks.

\section{Limitations}
Our work focuses on the TriviaQA and Natural Questions benchmarks, which include questions about popular entities in Wikipedia. As discussed in Appendix~\ref{app:popularity}, our approach simulates real-world common attacks which are the most frequent type of attacks. However, for entities that appear less often in the knowledge source (and are less likely to be attacked), our approach will not be as effective. 

We leave attacks on less-popularity entities to future work, as we focus on the most frequent and higher impact attacks, while also using datasets that are standard in existing literature, e.g. Natural Questions and TriviaQA. 

Our work shows the impact that disinformation attacks could have on Wikipedia and provides an initial attempt to help remedy those attacks. We note that our strategy does not have perfect accuracy and is still susceptible to attacks, e.g. if there is no correct information in any context to be found, it will be very difficult for ODQA systems to give the correct answer. We welcome additional research to improve the resistance of ODQA systems to disinformation attacks and will open-source our code and data to help others make progress in this area (including results from GPT-3).

\section*{Acknowledgements}
OW and NW are supported by the National Science Foundation Graduate Research Fellowship Program.

\bibliography{anthology,custom}

\appendix
\section{Realism of Proposed Setting}
\label{app:popularity}
We focus on data poisoning attacks to high to medium popularity entities, as included in TriviaQA and Natural Questions. 
But are such attacks realistic, and have they happened before?

Due to the way that search engines work, any data poisoning done at the time of indexing is able to effect system performance until the data is re-indexed. Thus, if one were to change a Wikipedia page (or a personal website that was included in an index) and that change was indexed, the data would be poisoned until re-indexing. 

As the people directing disinformation campaigns are likely motivated to attack well-known entities rather than unknown entities (for political or economic reasons), our proposed setting of defending against popular entities is well-motivated and is a serious problem affecting current production systems today.
(see \href{https://en.wikipedia.org/wiki/Vandalism_on_Wikipedia}{``Vandalism on Wikipedia"}). 
There have even been many high profile attacks on popular entities that have been reflected in production systems (this is not hypothetical). One such entity who has been frequently attacked is Donald Trump, whose Wikipedia page was changed to include critical text and inappropriate images, \textbf{returned by Siri to real user queries}. The Wikipedia page on vandalism includes many such examples of famous politicians, musicians, athletes and other popular entities being subject to attacks on Wikipedia that were propagated to users via Google or via various news outlets (e.g. Thomas Edison's page describing him as a "douchebag", famed swimmer Chad Le Clos's page edited to say he literally "died at the hands of Michael Phelps" when losing a race, etc.).

These attacks are just the tip of the iceberg for disinformation, as attacks to Wikipedia are the easiest to trace. Since production search engines index the web and then answer questions about them, any personal or company page can be used for attacks and are much less traceable (see this humorous attack to \href{https://twitter.com/mark_riedl/status/1637986261859442688?s=20}{Bing Chat about Mark Reidl}, done in jest to illustrate the potential for attacks).

\section{Hyperparameters}
\label{app:hyperparameters}
For all our experiments we use a cluster of V100 GPUs, with each job running on a 4 to 8 GPU node and taking approximately 12-24 hours depending on the model. We use the models as provided by the original authors with default retriever hyperparameters. We use \textsc{Atlas}'s XL version. We use Vicuna v1.5 on 1 A100 40GB GPU for 3 hours for the open-source experiment in Appendix~\ref{app:llama}.

Following previous work in question answering, we report Exact Match (EM) in all of our experiments. We take the data from \citet{longpre2021entity} and split into equal dev and test sets. We use the dev set to tune the CAR method's hyperparameters and use $K=5$ for our experiments.

Along with the \textit{New C} and \textit{Original C} options, the \textit{redundancy} and \textit{majority vote} methods also have hyperparameters for using either the augmented questions or the original question for the final prediction (after generating and searching for new contexts). Our tuning on the dev set indicated that using the original question and the new contexts from searching with the augmented question provides slightly higher performance (which makes sense, since the original question is the most important to answer). Thus, the process is first generating augmented questions, then searching with those, then doing inference with the original questions and the newly retrieved contexts (and finally CAR, if using the redundancy method).

\section{More Related Work}
\label{app:related}
As a larger section of related work did not have space in the main paper, we include more related work here.

\paragraph{Data Poisoning Attacks}
Data poisoning attacks in NLP have a long history, with several prominent works appearing in recent years including \cite{Wallace2019Triggers,Wallace2020Stealing,schwarzschild2021just} focusing on various NLP tasks such as machine translation, language modeling, etc. However, in the question answering space most adversarial work is focused on making harder questions, rather than simulating a real attack \cite{wallace2019trick,lee2019domain}. Those that do focus on human attacks focus on the machine reading setting \cite{bartolo2021improving}.

As mentioned in the main text, a nascent line of work has focused on knowledge conflicts in open-domain question answering \cite{Chen2022RichKS,longpre2021entity}. These works' main motivation is to explore how ODQA models operate under the influence of conflicts, mostly in the context of non-parametric vs parametric knowledge. We extend these works by using their methods as simulated attacks on a knowledge source and proposing efforts to defend against these attacks.

\paragraph{Open-Domain Question Answering}
Our work builds off of recent advances in ODQA, such as using Fusion-in-Decoder \cite{Izacard2021LeveragingPR}.
Other work such as DPR \cite{karpukhin2020dense} showed promising results but has been improved upon by models that encode a large number of contexts into a single reader model. We note there exists an emerging line of work that uses LLMs for ODQA without using a retriever \cite{zhou2023context,weller2023according}, however, our approach relies on the redundancy in the retriever to defend against disinformation attacks; we leave exploring other settings to future work.

\paragraph{Query Augmentation}
Query augmentation is a traditional information retrieval technique to augment a given query to find a better set of documents \cite{singhal2001modern,carpineto2012survey}. In classical terms, the strategy is usually to expand the query, spelling out acronyms or adding synonyms. Recently, work has begun to use neural models to generate these expansions \cite{wang2021pseudo,claveau2021neural,jagerman2023query,weller2023generative}, despite retriever's lack of understanding of some terms \cite{weller2023nevir}. In our work, we use a similar strategy to create new queries that will gather a diverse set of passages. 

\paragraph{Confidence and Calibration of QA}
Many works have focused on calibrating QA models so that they correctly reflect probabilities that equal their actual correct answer rate \cite{clark2017simple,kamath2020selective,si2022revisiting,jiang2021can}. Our proposed confidence method is similar in that it measures when the model will be more likely to be correct, however, it does not do calibration in the sense of calibrated probabilities, instead giving a single value of ``confident" or ``not confident."

Answer redundancy has been studied before in other NLP contexts, such as \citet{downey2006probabilistic}  in the information extraction task. We apply a similar intuition of answer redundancy to the novel context of document inputs for open-domain question answering.

\begin{figure}[t!]
     \centering
     \includegraphics[scale=.65,trim={3cm 0cm 2cm 0cm}]{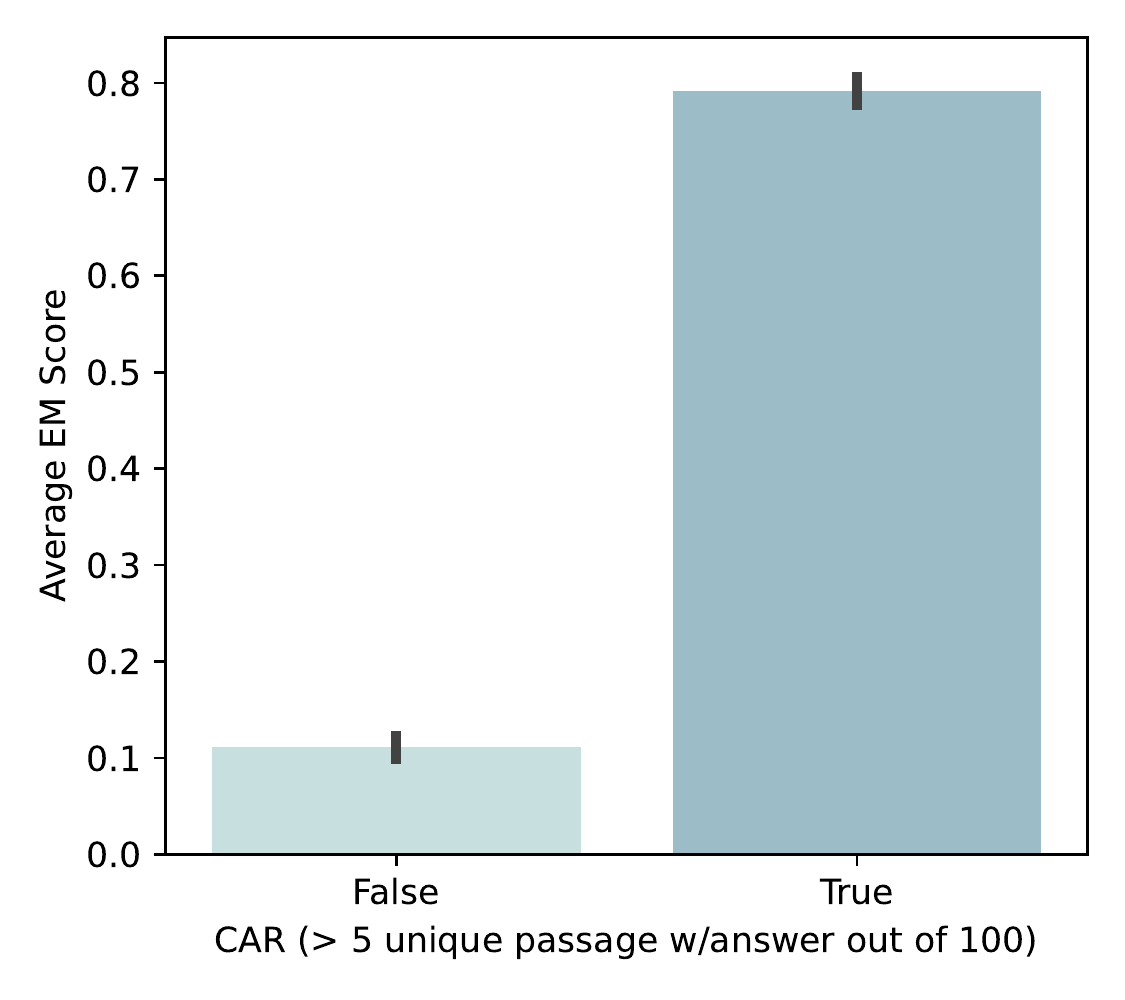}
     \caption{An ablation on Confidence from Answer Redundancy (CAR) compared to their exact match scores on the NQ 1-article poisoned setting. Those in the True bar have greater than 5 unique passages that contain the predicted answer string.}
     \label{fig:car}
\end{figure}

\begin{figure*}[t!]
     \centering
     \includegraphics[scale=.65,trim={2cm 0cm 0cm 0cm}]{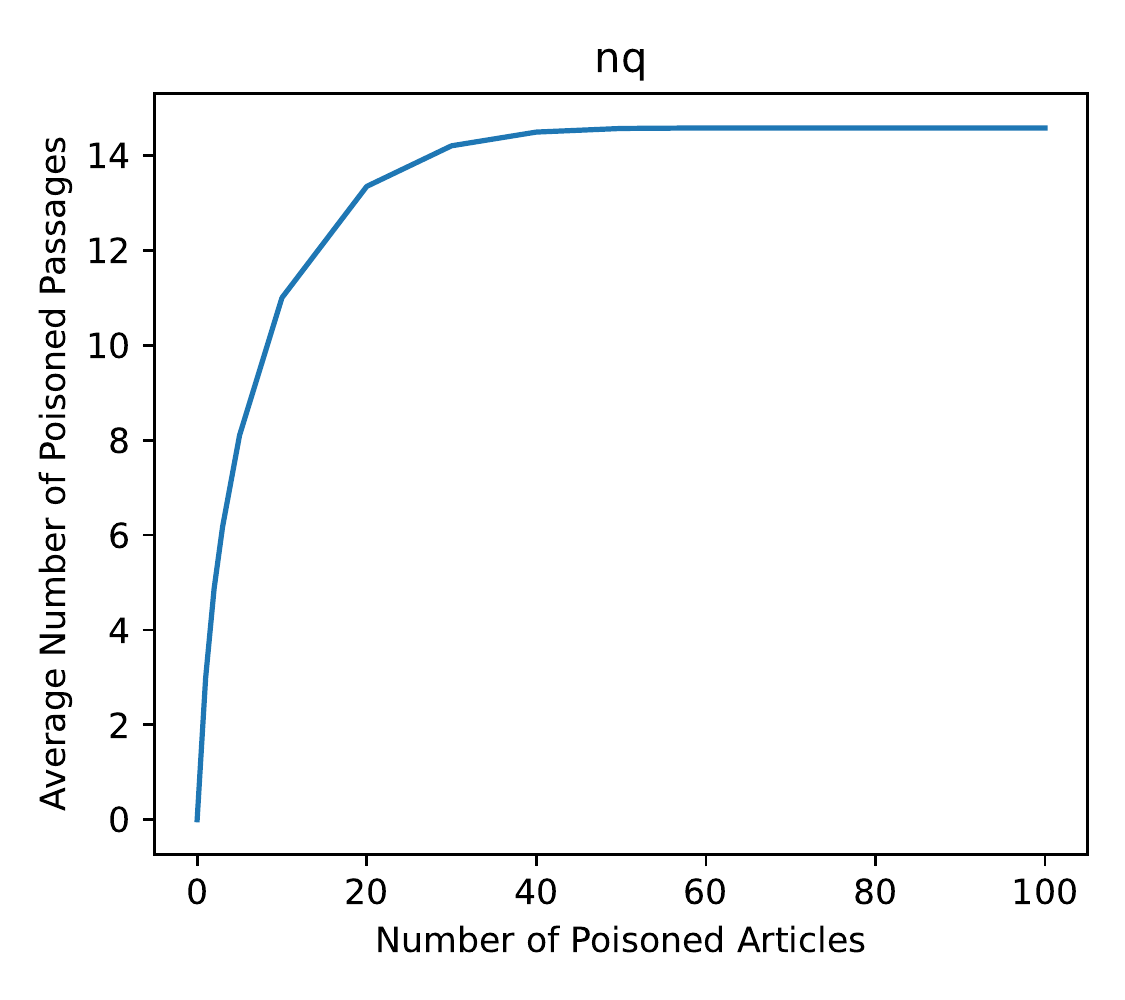}
      \includegraphics[scale=.65,trim={0cm 0cm 2cm 0cm}]{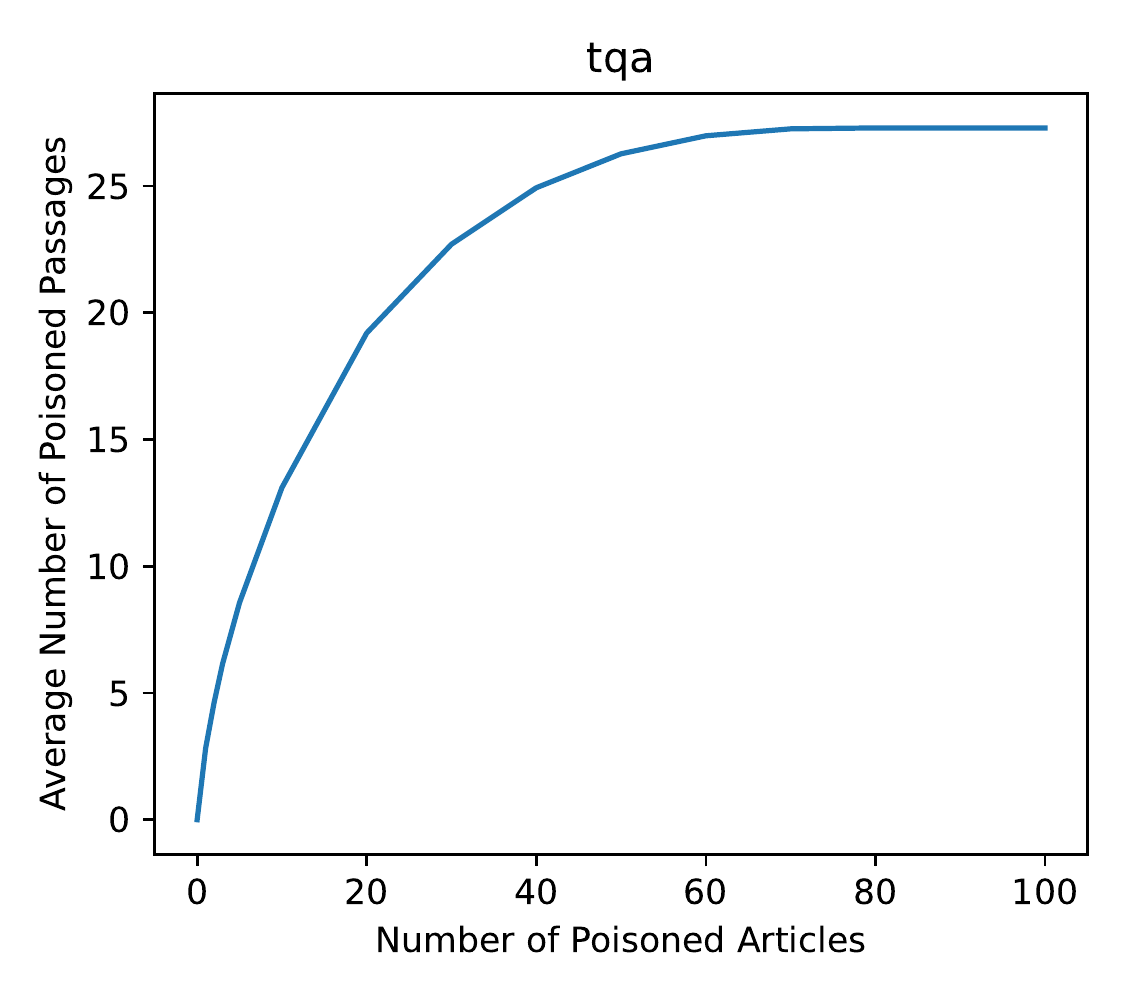}
     \caption{The number of poisoned \textit{passages} given the article poisoning level. Notice that TriviaQA (\textit{tqa}, right) has more passages to poison and a more gradual slope of poisoning than Natural Questions (\textit{nq}, left).}
     \label{fig:num_poisoned}
\end{figure*}

\section{Alternate Poisoning Attacks}
\label{app:poisoned_contexts}
In the main section of the paper, we used poisoning attacks based on articles. However, one could attack a system directly by going after its retrieved results, either randomly poisoning $N$\% or poisoning the top $N$\%. We note that we tried both settings and found similar results, with the main difference that model performance declines slower (as randomly picking contexts to poison is less likely to impact the model until higher levels of poisoning).

\section{Number of Augmented Queries}
\label{app:num_queries}
In Figure~\ref{fig:number_queries} we see the results for how the number of augmented queries affects performance. Overall, one query provides strong performance (above the baseline original performance at 17.5\% EM) and multiple questions continue to show gains. We note that this figure uses Natural Questions and the 5-article poisoning setting with FiD, but other settings showed roughly the same results. As including more queries only seems to increase the score, it's possible that generating more than 10 augmented queries would show even better results.

\section{Why is performance not 0\% at 100 poisoned documents?}
\label{app:not_zero}
To explore this question, we conducted a manual analysis of 20 pairs of question and 100 document passages on TriviaQA using FiD. We found that 65\% of cases were due to the model's parametric knowledge, as there was no such answer string in the input text. However, the answer was generally very obvious, like ``In which country is Dubrovnik?" which is generally easier for the model to predict (e.g. ``Croatia"). In 35\% of cases there was a missing alias from the answer string set, such as ``What dance craze was named after a city in South Carolina?" with an answer string set of ``Charleston rhythm", ``Charleston (dance)", ``Charleston (dance move)", ``Charleston dance", and ``The Charleston". FiD predicted ``Charleston" from the text, since ``Charleston" was not in the answer string set so it was not poisoned in the text. Future work on data poisoning could improve on this category by developing more robust poisoning techniques to aliases.

\section{Number of Poisoned Passages}
\label{app:num_poisoned}
In our experiments, we poisoned at the article level, as an attacker might do to a specific entity. However, each Wikipedia article corresponds to more than one \textit{passage} which are what is used for retrieval. When we poison at the article level we poison all passages in the article, so oftentimes many passages are poisoned even when poisoning one article. Furthermore, passages can only be poisoned if the answer is present in the passage (and thus available to be replaced).

How many passages are poisoned at each article-poisoning level? \autoref{fig:num_poisoned} answers this question and shows the number of poisoned passages vs the article-poisoning level. We find that the number of articles poisoned is much higher on TriviaQA, which means that TriviaQA had a much higher number of passages with the answer to begin with.

\begin{figure*}
\centering
\begin{subfigure}{.5\textwidth}
  \centering
  \includegraphics[width=.75\linewidth,trim={2cm 0.5cm 1cm 0cm}]{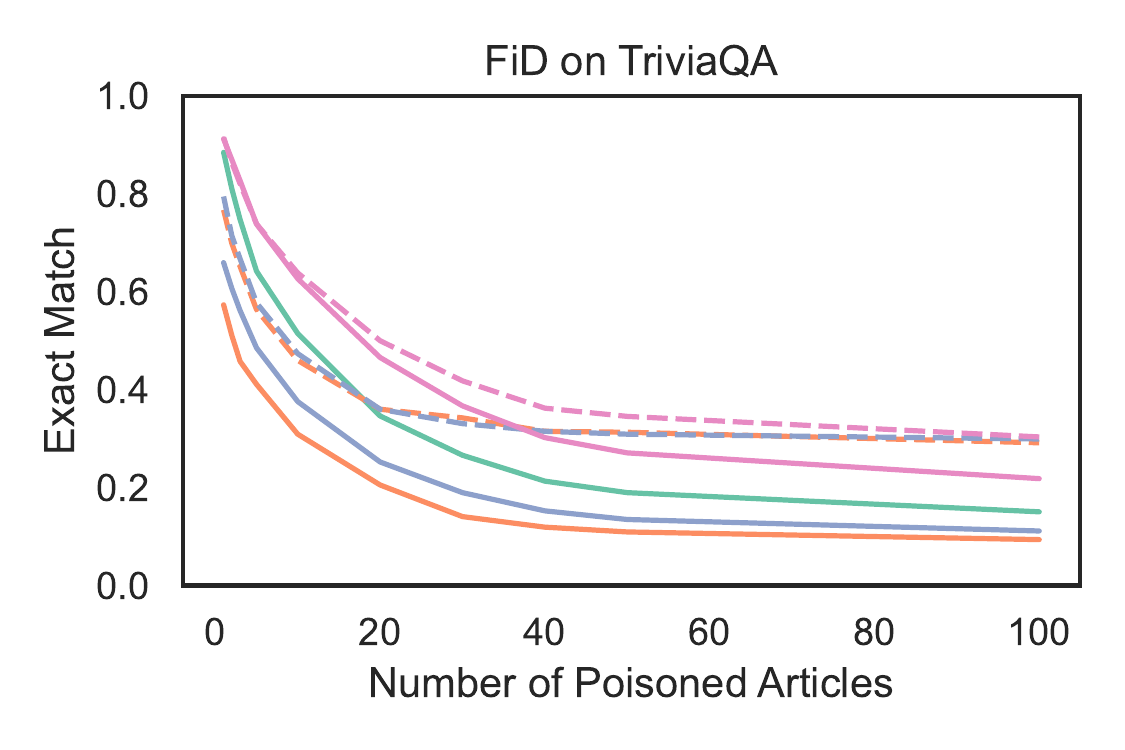}
  \label{fig:sub1}
\end{subfigure}%
\begin{subfigure}{.5\textwidth}
  \centering
  \includegraphics[width=.75\linewidth,trim={2cm 0.5cm 1cm 0cm}]{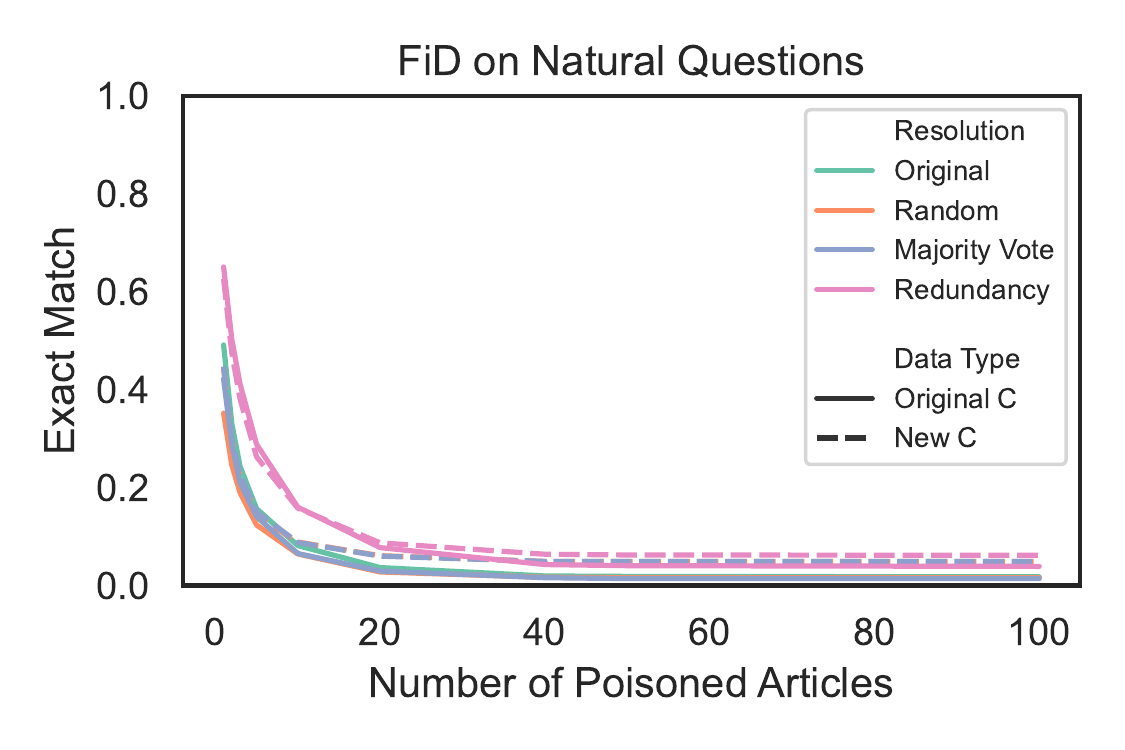}
  \label{fig:sub2}
\end{subfigure}
\caption{Main results showing the effect of data poisoning and various defense strategies on TriviaQA and Natural Questions using FiD as the retrieval augmented model and \textbf{Vicuna v1.5} as the question augmentation model. Q and C stand for question and context respectively. We see that open-source models can still provide similar gains.}
\label{fig:main_llama}
\end{figure*}

\section{Confidence from Answer Redundancy}
We compare the confidence from answer redundancy (CAR) to the actual exact match score (using the 1-article poisoning setting on Natural Questions) to show the effectiveness of this heuristic. In Figure~\ref{fig:car} we see the large gap between queries that do not meet CAR and those that do (around 65\% absolute exact match). Error bars indicate a 95\% confidence interval.

\section{Alternate Answer Resolution Strategies}
\label{app:more_baselines}
Due to space and clarity for figures, we do not include all possible answer resolution strategies in the main figures. Some potential alternate resolution stratgies we tried included: 
\begin{itemize}
    \item Using the new augmented questions with CAR alone, without using them as a backup for the original question. This is equivalent to the majority vote method but using CAR to filter the question that get to vote. Although this method performed well it consistently underperformed our \textit{redundancy} method and thus we do not include it
    \item Using a majority vote over both the original question's prediction \textit{and} and augmented question's predictions. This performed nearly identically to the standard majority vote method, hence we leave it out for clarity.
    \item Taking the difference between the the CAR values of the original and augmented questions. This again greatly underperformed the \textit{redundancy} method and is therefore not included
\end{itemize}

We encourage others who have new ideas for answer resolution strategies to use our code as a start to develop their method.

\section{Compute Cost of our Proposed Method}
\label{app:compute}
Our method requires the addition of 1 call to GPT-3's API (or the use of Llama 2, see Appendix~\ref{app:llama}) which generates the N augmented questions with one call, and N instances of additional search and inferences of the ODQA model. 

\paragraph{Augmented Query Generation} As GPT-3 and other large language models become more available and cheaper (as they have already started to be, with many works speeding up inference using models like Llama) this will become cheaper to do with time. The cost of one query to ChatGPT for example (of which our method uses approximately 100 tokens) is roughly \$0.0002 USD, which is remarkably affordable.

\paragraph{Retrieval} The retrieval computation cost is negligible in comparison, as modern retrieval takes milliseconds with different indexing and quantitation costs. 

\paragraph{Retrieval-Augmented Generation} The other major computational cost of our method is the retrieval augmented generation inference. However, as mentioned  \autoref{app:num_queries}, this can be reduced to only one inference and still see large gains.

\begin{figure*}[t!]
    \centering
    \vspace*{0.25em}
     \large{Example (1)} \\
     \begin{tabular}{l|p{11.2cm}}
      \toprule
     \textbf{\small{Original Question}}    &  \small{In 2010 British ex-soldier Ed Stafford became the first person (ever known) to walk the entire length of what river?} \\
     \textbf{\small{Original Doc Rank \#1}}    & \small{Ed Stafford Edward James Stafford FRSGS, known as Ed Stafford ... for being the first human ever to walk the length of the \textcolor{red}{Panthers River} ...} \\
     \textbf{\small{Original Prediction}} & \small{\textcolor{red}{Panthers}} \\
     \cmidrule(lr){1-1} \cmidrule(lr){2-2}
     \textbf{\small{Augmented Question}} &  \small{What river did Ed Stafford cross in 2010?} \\ 
     \textbf{\small{New Doc Rank \#1}} & \small{... the two men, Stafford and Sanchez Rivera, walked for a further two years before reaching the mouth of the \textcolor{blue}{Amazon River} on 9 August 2010 ...} \\ 
     \textbf{\small{New Answer (correct)}} & \small{\textcolor{blue}{Amazon}} \\
     \bottomrule
     \end{tabular} 
    \newline
    \vspace*{1.25em}
    \newline
    \large{Example (2)} \\
    \begin{tabular}{l|p{11.2cm}}
         \toprule
     \textbf{\small{Original Question}} & \small{What is the first name of Irish singer Van Morrison?} \\

    \textbf{\small{Original Doc Rank \#1}} &  \small{\textcolor{red}{Boutros Ghali Ivan "Van" Morrison} was born on 31 August 1945, at 125 Hyndford Street, Bloomfield, Belfast, Northern Ireland,  ...} \\
    
    \textbf{\small{Original Prediction}} & \small{\textcolor{red}{Boutros}} \\
    \cmidrule(lr){1-1} \cmidrule(lr){2-2}
    \textbf{\small{Augmented Question}} & \small{Can you give me any information about the first name of Irish singer Van Morrison?} \\
    \textbf{\small{New Doc Rank \#1}}& .\small{.. book also contains a complete discography of Van Morrison\'s work. Turner describes Van Morrison\'s early life as \textcolor{blue}{George Ivan Morrison} on Hyndford Street in Belfast ...} \\
    \textbf{\small{New Answer (correct)}} & \small{\textcolor{blue}{George}} \\
     \bottomrule
    \end{tabular}
    
\caption{Case study illustrating differences in QA predictions using original and augmented questions. We show incorrect answers/predictions in \textcolor{red}{red} and correct answers/predictions in \textcolor{blue}{blue}. These examples show how the augmentation helps: in (1) the augmented question focuses more on the river than the person (e.g. by removing personal details) and by re-weighting query terms is able to correctly rank the Wikipedia page for ``Walking the Amazon" higher. In (2) the augmented question is more vague (``information" rather than ``first name"), allowing it to rank the Wikipedia page for his biography higher than his poisoned personal page.}
\label{fig:case-study}
\end{figure*}

\paragraph{Overall} Our method is bounded by the call to a language model for generation of the augmented questions and by 1 or more calls to a retrieval augmented model. However, these costs are still cheap and used frequently: e.g. one call at inference time to a model like ChatGPT is relatively minor and is done by a large number of research and industry applications. Further, as time progresses these calls will get cheaper and quicker.

\section{Open-Source LM Generation}
\label{app:llama}
We also show that our method can use an open-source language model like Llama 2 \cite{touvron2023llama} and work similarly. In Tables~\ref{tab:llama_tqa_fid} and \ref{tab:llama_nq_fid} and Figure~\ref{fig:main_llama} we show results for FiD on TriviaQA and Natural Questions using Llama 2 generations from Vicuna v1.5 7B \cite{zheng2023judging}.

We find that results are comparable to those with GPT-3, and in some cases even slightly outperforms the comparable GPT-3 version. For example, in the 1\% article poisoning case Vicuna v1.5 scores 91.5\% with redundancy while GPT-3 score 90.8\%, Table~\ref{tab:llama_tqa_fid} vs Table~\ref{tab:full_tqa_fid}. 

Overall, we see that open-source models can comparably be used in place of closed models like GPT-3 for this task.

\section{Relation to Robustness under Shift}
One common type of evaluation in ODQA is its ability to withstand adversarial attacks that test robustness (such as paraphrases or distracting sentences that are superfluous), e.g. \citet{yang2018hotpotqa,gan2019improving,Yoran2023MakingRL}. However, our work focuses on intentional disinformation attacks, where the facts in the documents have been changed. Although these two evaluation settings have surface similarities, the crucial difference is that when adding distracting sentences or paraphrases the core facts still exist in the documents, with no contradictions between documents.  In our experimental setting however, some or all the needed facts have been deliberately changed. Thus, techniques for handling distracting sentences (such as improved filtered) or paraphrases (training with paraphrases) are not relevant to our setting, as the underlying problem requires new solutions that can deal with incorrect and/or conflicting facts.

\section{Case Studies and Examples}
\label{app:case_study}
We show two case studies here that illustrate how our method works. We randomly select an instance where our method outperforms the baseline approach. 

\subsection{Case 1} We see that the new query successfully re-weighted terms such that it was able to rank the new document \#1 and get the correct answer. Note that the New Rank 1 document was from the ``Walking the Amazon" page, which was not poisoned, while the Original Rank 1 document was from the poisoned Ed Stafford page.  Note that both of these documents were in the top three for each query, but the relative position change was able to help the retrieval-augmented model find the correct answer. 

\subsection{Case 2} In this example we see that searching for ``information" allowed the query to find the Wikipedia page which described Turner's book about Van Morrison, which contained the correct answer as opposed to the poisoned content. Having the correct answer in the top context allowed the model to correctly choose it over the disinformation.

\subsection{Case Study Conclusion} 
From a qualitative analysis, we find that our method predicted correctly on some instances because it changed the relative position of the retrieved documents, due to word changes in the query. Others were correctly predicted by surfacing new information that was not in the top ranked contexts before.  Our method thus makes relatively simple changes that, overall, provides strong gains and is easy to implement with any retrieval augmented system.

\section{Table Versions of Plots}
\label{app:tables}
We also show Table versions of the main plots for ease of viewing: Table~\ref{tab:full_tqa_fid} for TQA and FID, Table~\ref{tab:full_nq_fid} for NQ and FiD, Table~\ref{tab:full_nq_atlas} for NQ and \textsc{Atlas}, and Table~\ref{tab:full_tqa_atlas} for TQA and \textsc{Atlas}.

\section{F1 vs EM}
\label{app:f1}
It is common in previous work on NQ and TQA to report only EM.  However, we also include tables with F1 to illustrate that the results are the same, just slightly higher. As the differences are very minor and the trends remain the same, we three examples using TQA: Llama with FiD (Table~\ref{tab:llama_tqa_fid_f1}) and GPT-3 for \textsc{Atlas}  and FiD in Tables~\ref{tab:full_tqa_atlas_f1} and \ref{tab:full_tqa_fid_f1}.

\begin{table*}
\centering
\begin{tabular}{ll|rrrrrrrrr}
\toprule
& & \multicolumn{9}{c}{EM Scores at \# of Poisoned Articles} \\
Context Type & Resolution & 1 & 2 & 3 & 5 & 10 & 20 & 40 & 50 & 100 \\ 
\midrule
 \multirow{4}{*}{Original C} & Majority Vote & 55.9 & 44.5 & 38.2 & 28.4 & 19.3 & 10.8 &  6.7 &  5.9 &  5.6 \\
  &      Original & 69.4 & 55.4 & 47.0 & 35.9 & 24.3 & 14.4 &  9.7 &  8.0 &  7.8 \\
 &        Random & 44.1 & 36.0 & 31.0 & 24.0 & 15.3 &  9.5 &  5.9 &  5.1 &  4.1 \\
  &    Redundancy & 78.5 & 69.7 & 62.2 & 52.3 & 37.6 & 22.3 & 15.0 & 12.1 & 11.8 \\
 \midrule
\multirow{3}{*}{New C} & Majority Vote & 65.1 & 52.6 & 44.8 & 34.2 & 22.9 & 16.5 & 12.9 & 12.3 & 12.2 \\
 &        Random & 64.3 & 51.0 & 42.7 & 31.7 & 21.1 & 15.9 & 12.3 & 11.9 & 11.1 \\
 &    Redundancy & 78.1 & 67.3 & 60.2 & 50.5 & 37.7 & 25.5 & 19.8 & 17.6 & 17.5 \\
\bottomrule
\end{tabular}
\caption{Full results for NQ with ATLAS on varying amounts of article poisoning. Results in EM.}
\label{tab:full_nq_atlas}
\end{table*}

\begin{table*}
\centering
\begin{tabular}{ll|rrrrrrrrrr}
\toprule
& & \multicolumn{9}{c}{EM Scores at \# of Poisoned Articles} \\
Context Type & Resolution & 1 & 2 & 3 & 5 & 10 & 20 & 40 & 50 & 100 \\ 
\midrule
 \multirow{3}{*}{Original C} & Majority Vote & 61.5 & 55.2 & 51.9 & 45.2 & 38.1 & 30.5 & 25.1 & 25.1 & 22.2 \\
&      Original & 91.6 & 81.6 & 75.3 & 66.9 & 53.6 & 43.9 & 32.6 & 32.6 & 29.3 \\
&        Random & 47.7 & 43.1 & 40.6 & 35.1 & 28.0 & 21.8 & 17.6 & 16.7 & 17.6 \\
 &    Redundancy & 94.5 & 88.0 & 82.9 & 78.4 & 66.5 & 57.6 & 43.9 & 43.6 & 39.9 \\
 \midrule
\multirow{3}{*}{New C} & Majority Vote & 86.2 & 79.1 & 73.6 & 65.3 & 55.2 & 48.1 & 43.5 & 42.7 & 42.3 \\
    & Random & 85.4 & 77.8 & 72.8 & 64.0 & 55.2 & 48.5 & 44.4 & 43.9 & 43.5 \\
   & Redundancy & 95.1 & 88.5 & 82.6 & 77.7 & 68.5 & 59.9 & 51.4 & 50.3 & 47.2 \\
\bottomrule
\end{tabular}
\caption{Full results for TQA with ATLAS on varying amounts of article poisoning. Results in EM.}
\label{tab:full_tqa_atlas}
\end{table*}

\begin{table*}
\centering
\begin{tabular}{ll|rrrrrrrrrr}
\toprule
& & \multicolumn{9}{c}{EM Scores at \# of Poisoned Articles} \\
Context Type & Resolution & 1 & 2 & 3 & 5 & 10 & 20 & 40 & 50 & 100 \\ 
\midrule
 \multirow{3}{*}{Original C} & Majority Vote & 66.8 & 59.7 & 54.2 & 46.2 & 36.8 & 24.9 & 15.4 & 13.0 & 11.5 \\
 &      Original & 87.0 & 79.4 & 72.3 & 61.7 & 49.4 & 31.6 & 20.2 & 17.4 & 13.8 \\
&        Random & 53.0 & 47.8 & 43.1 & 38.3 & 30.8 & 18.6 & 13.0 & 10.7 &  8.3 \\
&    Redundancy & 89.7 & 85.0 & 79.4 & 70.0 & 60.6 & 42.1 & 26.9 & 24.0 & 19.2 \\
 \midrule
\multirow{3}{*}{New C}  & Majority Vote & 83.8 & 76.7 & 70.0 & 58.9 & 48.2 & 35.2 & 31.6 & 30.8 & 28.9 \\
&        Random & 79.4 & 71.9 & 63.6 & 53.0 & 42.7 & 34.4 & 31.2 & 30.0 & 28.9 \\
  &    Redundancy & 90.8 & 86.7 & 81.8 & 71.1 & 62.8 & 47.5 & 37.5 & 35.9 & 33.2 \\
\bottomrule
\end{tabular}
\caption{Full results for TQA with FiD on varying amounts of article poisoning. Results in EM.}
\label{tab:full_tqa_fid}
\end{table*}

\begin{table*}
\centering
\begin{tabular}{ll|rrrrrrrrrr}
\toprule
& & \multicolumn{9}{c}{EM Scores at \# of Poisoned Articles} \\
Context Type & Resolution & 1 & 2 & 3 & 5 & 10 & 20 & 40 & 50 & 100 \\ 
\midrule
 \multirow{3}{*}{Original C} & Majority Vote & 41.6 & 28.0 & 22.0 & 13.4 &  6.2 & 3.2 & 1.6 & 1.4 &  1.4 \\
 &      Original & 50.1 & 33.0 & 25.7 & 17.5 &  9.3 & 4.8 & 3.0 & 2.8 &  2.6 \\
 &        Random & 29.6 & 19.0 & 14.1 &  9.7 &  4.6 & 1.9 & 1.4 & 1.4 &  1.2 \\
&    Redundancy & 65.3 & 50.2 & 41.6 & 29.3 & 16.8 & 8.7 & 5.0 & 4.7 &  4.4 \\
 \midrule
\multirow{3}{*}{New C}  & Majority Vote & 49.4 & 33.2 & 25.4 & 18.0 & 11.1 &  7.9 & 7.4 & 7.2 &  7.2 \\
&        Random & 46.9 & 31.4 & 23.5 & 16.8 & 10.8 &  8.5 & 7.6 & 7.6 &  6.7 \\
 &    Redundancy & 63.9 & 47.0 & 39.8 & 29.7 & 18.5 & 11.7 & 9.5 & 9.2 &  9.1 \\
\bottomrule
\end{tabular}
\caption{Full results for NQ with FiD on varying amounts of article poisoning. Results in EM.}
\label{tab:full_nq_fid}
\end{table*}

\begin{table*}
\centering
\begin{tabular}{ll|rrrrrrrrrr}
\toprule
& & \multicolumn{8}{c}{EM Scores at \# of Poisoned Articles} \\
Context Type & Resolution & 1 & 2 & 3 &  5 & 10 & 20 & 40 & 50 & 100 \\ 
\midrule
 \multirow{4}{*}{Original C} & Majority Vote & 65.9 & 60.7 & 56.2 & 48.5 & 37.6 & 25.2 & 15.3 & 13.5 & 11.2 \\
 &      Original & 88.5 & 81.0 & 74.8 & 64.2 & 51.5 & 34.6 & 21.3 & 19.0 & 15.1 \\
 &        Random & 57.3 & 51.1 & 45.8 & 41.1 & 30.9 & 20.5 & 11.9 & 11.0 &  9.4 \\
 &    Redundancy & 91.2 & 87.0 & 82.7 & 73.8 & 62.7 & 46.6 & 30.2 & 27.1 & 21.8 \\
 \midrule
\multirow{3}{*}{New C}  & Majority Vote & 79.5 & 71.4 & 66.7 & 57.7 & 47.4 & 36.0 & 31.5 & 30.9 & 29.9 \\
      &        Random & 76.7 & 69.9 & 65.2 & 56.4 & 46.0 & 36.0 & 31.5 & 31.3 & 29.2 \\
       &    Redundancy & 91.5 & 86.3 & 82.0 & 73.8 & 63.8 & 50.0 & 36.2 & 34.6 & 30.4 \\
\bottomrule
\end{tabular}
\caption{Full results for TQA with FiD with Llama 2 Vicuna v1.5 generations on varying amounts of article poisoning. Results in EM. Note that results are comparable to GPT-3 DaVinci used in the main text and in Table~\ref{tab:full_tqa_fid}.}
\label{tab:llama_tqa_fid}
\end{table*}

\begin{table*}
\centering
\begin{tabular}{ll|rrrrrrrrrr}
\toprule
& & \multicolumn{8}{c}{EM Scores at \# of Poisoned Articles} \\
Context Type & Resolution & 1 & 2 & 3 &  5 & 10 & 20 & 40 & 50 & 100 \\ 
\midrule
 \multirow{4}{*}{Original C} & Majority Vote & 42.2 & 28.8 & 21.0 & 14.1 &  6.6 & 3.0 & 1.6 & 1.4 &  1.4 \\
&      Original & 49.1 & 33.1 & 24.3 & 15.7 &  8.2 & 3.7 & 2.0 & 1.9 &  1.8 \\
 &        Random & 35.2 & 24.6 & 19.0 & 12.4 &  6.5 & 2.8 & 1.7 & 1.6 &  1.6 \\
&    Redundancy & 65.1 & 50.3 & 41.2 & 28.9 & 16.0 & 7.8 & 4.3 & 4.1 &  3.9 \\
 \midrule
\multirow{3}{*}{New C}  & Majority Vote & 44.7 & 30.0 & 22.6 & 15.1 &  8.8 & 6.0 & 4.9 & 4.9 &  4.9 \\
&        Random & 44.3 & 29.6 & 22.3 & 14.4 &  8.9 & 6.1 & 4.9 & 4.9 &  4.9 \\
 &    Redundancy & 62.6 & 47.4 & 38.1 & 26.3 & 15.8 & 8.7 & 6.4 & 6.3 &  6.2 \\
\bottomrule
\end{tabular}
\caption{Full results for NQ with FiD with Llama 2 Vicuna v1.5 generations on varying amounts of article poisoning. Results in EM. Note that results are comparable to GPT-3 DaVinci used in the main text and in Table~\ref{tab:full_nq_fid}.}
\label{tab:llama_nq_fid}
\end{table*}

\begin{table*}
\centering
\begin{tabular}{ll|rrrrrrrrrr}
\toprule
& & \multicolumn{9}{c}{F1 Scores at \# of Poisoned Articles} \\
Context Type & Resolution & 1 & 2 & 3 & 5 & 10 & 20 & 40 & 50 & 100 \\ 
\midrule
 \multirow{3}{*}{Original C} & Majority Vote & 64.9 & 59.2 & 56.1 & 49.1 & 42.0 & 34.7 & 29.2 & 29.0 & 26.4 \\
            &      Original & 92.1 & 82.6 & 76.8 & 68.5 & 56.9 & 47.6 & 36.7 & 36.7 & 34.1 \\
            &        Random & 52.7 & 47.7 & 45.0 & 39.4 & 32.2 & 25.3 & 21.1 & 20.5 & 22.0 \\
            &    Redundancy & 94.6 & 88.4 & 83.8 & 79.1 & 68.4 & 59.1 & 46.0 & 45.7 & 42.2 \\
 \midrule
\multirow{3}{*}{New C} & Majority Vote & 87.9 & 80.7 & 75.3 & 67.2 & 58.8 & 51.4 & 47.3 & 46.5 & 46.1 \\
            &        Random & 86.9 & 79.8 & 74.3 & 66.0 & 58.7 & 51.9 & 47.8 & 47.6 & 47.4 \\
            &    Redundancy & 95.4 & 89.0 & 83.7 & 78.5 & 70.6 & 62.4 & 54.2 & 53.2 & 50.5 \\
\bottomrule
\end{tabular}
\caption{Full results (in F1) for TQA with ATLAS on varying amounts of article poisoning. Results in F1.}
\label{tab:full_tqa_atlas_f1}
\end{table*}

\begin{table*}
\centering
\begin{tabular}{ll|rrrrrrrrrr}
\toprule
& & \multicolumn{9}{c}{F1 Scores at \# of Poisoned Articles} \\
Context Type & Resolution & 1 & 2 & 3 & 5 & 10 & 20 & 40 & 50 & 100 \\ 
\midrule
 \multirow{3}{*}{Original C} & Majority Vote & 68.7 & 61.4 & 55.7 & 48.0 & 39.0 & 27.5 & 17.7 & 15.5 & 13.9 \\
            &      Original & 87.2 & 80.4 & 73.4 & 63.4 & 51.3 & 34.1 & 22.8 & 20.5 & 17.2 \\
            &        Random & 56.0 & 50.7 & 45.4 & 40.2 & 32.6 & 20.8 & 14.9 & 13.1 & 11.0 \\
            &    Redundancy & 90.0 & 85.4 & 79.8 & 71.0 & 61.3 & 43.7 & 28.5 & 26.0 & 21.4 \\
 \midrule
\multirow{3}{*}{New C}   & Majority Vote & 84.2 & 77.2 & 70.5 & 59.6 & 49.9 & 36.8 & 33.4 & 32.6 & 30.7 \\
            &        Random & 79.7 & 72.3 & 64.0 & 53.8 & 43.7 & 35.5 & 32.4 & 31.4 & 30.2 \\
            &    Redundancy & 90.9 & 87.0 & 82.1 & 72.1 & 64.0 & 49.0 & 38.6 & 37.4 & 34.8 \\
\bottomrule
\end{tabular}
\caption{Full results (in F1) for TQA with FiD on varying amounts of article poisoning. Results in F1.}
\label{tab:full_tqa_fid_f1}
\end{table*}

\begin{table*}
\centering
\begin{tabular}{ll|rrrrrrrrrr}
\toprule
& & \multicolumn{8}{c}{F1 Scores at \# of Poisoned Articles} \\
Context Type & Resolution & 1 & 2 & 3 &  5 & 10 & 20 & 40 & 50 & 100 \\ 
\midrule
 \multirow{4}{*}{Original C} & Majority Vote & 67.7 & 62.3 & 57.9 & 50.5 & 39.6 & 27.5 & 17.3 & 15.6 & 13.3 \\
            &      Original & 88.6 & 81.3 & 75.2 & 65.2 & 52.8 & 37.1 & 23.7 & 21.5 & 17.6 \\
            &        Random & 59.2 & 52.9 & 47.9 & 43.3 & 33.2 & 23.2 & 14.2 & 13.3 & 11.8 \\
            &    Redundancy & 91.3 & 87.1 & 82.9 & 74.5 & 63.4 & 47.9 & 31.1 & 28.4 & 23.2 \\
 \midrule
\multirow{3}{*}{New C}  & Majority Vote & 79.7 & 71.8 & 67.3 & 58.7 & 48.5 & 37.5 & 33.2 & 32.6 & 31.7 \\
            &        Random & 76.9 & 70.3 & 65.8 & 57.3 & 47.2 & 37.4 & 33.2 & 33.0 & 31.3 \\
            &    Redundancy & 91.6 & 86.5 & 82.3 & 74.6 & 64.7 & 51.8 & 37.6 & 36.4 & 32.3 \\
\bottomrule
\end{tabular}
\caption{Full results (in F1) for TQA with FiD with Llama 2 Vicuna v1.5 generations on varying amounts of article poisoning. Results in EM. Note that results are comparable to GPT-3 DaVinci used in the main text and in Table~\ref{tab:full_tqa_fid}.}
\label{tab:llama_tqa_fid_f1}
\end{table*}

\end{document}